\documentclass[11pt,a4paper]{article}
\usepackage[hyperref]{emnlp2018}
\usepackage{times}
\usepackage{latexsym}
\usepackage{booktabs}
\usepackage{scalerel}
\usepackage{enumitem}
\usepackage{placeins}
\usepackage{url}
\usepackage{amsmath} 
\usepackage{framed}
\usepackage[table]{xcolor}

\aclfinalcopy % Uncomment this line for the final submission
\def\aclpaperid{***} %  Enter the acl Paper ID here

\newcommand{\notcopied}[1]{{\textcolor{gray}{[#1]}}}
\newcommand{\squeezeup}{\vspace{-3.5mm}}
\def\textequal{\texttt{=}}

\title{Bottom-Up Abstractive Summarization}

\author{Sebastian Gehrmann \quad \quad \quad Yuntian Deng \quad \quad \quad Alexander M. Rush \\
  School of Engineering and Applied Sciences \\
  Harvard University\\
  {\tt \{gehrmann, dengyuntian, srush\}@seas.harvard.edu}}

\date{}
\usepackage[acronym,smallcaps,nowarn,section,nogroupskip,nonumberlist]{glossaries}
\glsdisablehyper{}
\newacronym{CNNDM}{CNN-DM}{CNN-Daily Mail}
\newacronym{ROUGE-1}{Rouge-1}{Rouge-1}
\newacronym{ROUGE-2}{Rouge-2}{Rouge-2}
\newacronym{ROUGE-L}{Rouge-L}{Rouge-L}
\begin{document}
\maketitle
\begin{abstract}
Neural network-based methods for abstractive summarization produce outputs that are more fluent than other techniques, but perform poorly at content selection. This work proposes a simple technique for addressing this issue: use a data-efficient content selector to 
over-determine phrases in a source document that should be part of the summary. We use this selector as a \textit{bottom-up} attention step to constrain the model to likely phrases. We show that this approach improves the ability to compress text, while still generating fluent summaries. This two-step process is both simpler and higher performing than other end-to-end content selection models, leading to significant improvements on ROUGE for both the  \acrshort{CNNDM} and NYT corpus. Furthermore, the content selector can be trained with as little as 1,000 sentences, making it easy to transfer a trained summarizer to a new domain.
\end{abstract}

\section{Introduction}

Text summarization systems aim to generate natural language summaries that compress the information in a longer text. Approaches 
using neural networks have shown promising results on this task with end-to-end models that encode a source document and then decode it into an abstractive summary. Current state-of-the-art neural abstractive summarization models combine extractive and abstractive techniques by using  pointer-generator style models which can copy words from the source document \citep{gu2016incorporating,see2017get}. 
These end-to-end models produce fluent abstractive summaries but have had mixed success in content selection, i.e. deciding what to summarize, compared to fully extractive models.

% In this work, we consider a class of models inspired by recent advances in computer vision that constrain the content that an abstractive summarization model can generate. These models are much more effective about deciding which phrases a model should include in a summary, without sacrificing the fluency advantages of recent summarization models. Furthermore, they require much less data to train and can be easily adapted to new data. 

\begin{figure}[t!]
\begin{framed}
\textbf{Source Document}

\small{german chancellor angela merkel \notcopied{did} not \notcopied{look} too pleased about the weather during her \notcopied{annual} easter holiday \notcopied{in italy.} as britain \notcopied{basks} in \notcopied{sunshine} and temperatures of up to 21c, mrs merkel and her husband\notcopied{, chemistry professor joachim sauer,} had to settle for a measly 12 degrees. the chancellor and her \notcopied{spouse} have been spending easter on the small island of ischia, near naples in the mediterranean for over a \notcopied{decade.} 

\notcopied{not so sunny:} angela merkel \notcopied{and} her husband\notcopied{, chemistry professor joachim sauer,} are spotted on their \notcopied{annual} easter trip to the island of ischia\notcopied{,} near naples\notcopied{. the} couple \notcopied{traditionally} spend their holiday at the five-star miramare spa hotel on the south of the island \notcopied{, which comes} with its own private beach \notcopied{, and balconies overlooking the} ocean \notcopied{.}...}

\textbf{Reference}

\small{
\setlist{nolistsep}
\begin{itemize}[align=left, leftmargin=*, noitemsep]
\item angela merkel and husband spotted while on italian island holiday. \\
$\ldots$ 
\end{itemize}}

\textbf{Baseline Approach}

\small{
\setlist{nolistsep}
\begin{itemize}[align=left, leftmargin=*, noitemsep]
\item angela merkel and her husband, chemistry professor joachim sauer, are spotted on their annual easter trip to the island of ischia, near naples. \\
$\ldots$ %the couple traditionally spend their holiday at the five-star miramare spa hotel on the south of the island. 
\end{itemize}}

\textbf{Bottom-Up Summarization}

\small{
\setlist{nolistsep}
\begin{itemize}[align=left, leftmargin=*, noitemsep]
\item angela merkel and her husband are spotted on their easter trip to the island of ischia, near naples. 
\\ $\ldots$ %the couple had to settle for a measly 12 degrees.
\end{itemize}
}
\end{framed}
\caption{Example of two sentence summaries with and without bottom-up attention. The model does not allow copying of words in \notcopied{gray}, although it can generate words. With bottom-up attention, we see more explicit sentence compression, while without it whole sentences are copied verbatim.}
\label{fig:ex}
\squeezeup
\end{figure}

There is an appeal to end-to-end models from a modeling perspective; however, there is evidence that when summarizing people follow a two-step approach of first selecting important phrases and then paraphrasing them~\citep{anderson1988teaching,jing1999decomposition}.
%This approach leads to differences in summary style, for example on the CNN-DM dataset, sentences in human-generated summaries are on average 10 words long, over 3 words shorter than sentences from abstractive neural models.
A similar argument has been made for image captioning. \citet{anderson2017bottom} develop a state-of-the-art model with a two-step approach that first pre-computes bounding boxes of segmented objects and then applies attention to these regions. 
This so-called \textit{bottom-up} attention is inspired by neuroscience research describing attention based on properties inherent to a stimulus~\citep{buschman2007top}. 

%We use a similar approach for summarization, and restrict copying to phrases that are identified to likely be part of a summary by a separate content selection model. %However, since local salience is not the best indicator for whether a clause is actually relevant to the overall text, the abstractive model still needs to be able to keep a macro representation of a summary~\citep{van1980macrostructures,seidlhofer1995approaches}. 

Motivated by this approach, we consider \textit{bottom-up} attention for neural abstractive summarization. Our approach first selects a selection mask for the source document and then constrains a standard neural model by this mask. This approach can better decide which phrases a model should include in a summary, without sacrificing the fluency advantages of neural abstractive summarizers. Furthermore, it requires much fewer data to train, which makes it more adaptable to new domains.\looseness=-1

Our full model incorporates a separate content selection system to decide on relevant aspects of the source document. 
We frame this selection task as a sequence-tagging problem, with the objective of identifying tokens from a document that are part of its summary. We show that a content selection model that builds on contextual word embeddings~\citep{peters2018deep} can identify correct tokens with a recall of over 60\%, and a precision of over 50\%.  To incorporate bottom-up attention into abstractive summarization models, we employ masking to constrain copying words to the selected parts of the text, which produces grammatical outputs. We additionally experiment with multiple methods to incorporate similar constraints into the training process of more complex end-to-end abstractive summarization models, either through multi-task learning or through directly incorporating a fully differentiable mask.

Our experiments compare bottom-up attention with several other state-of-the-art abstractive systems. 
Compared to our baseline models of \citet{see2017get} bottom-up attention leads to an improvement in \acrshort{ROUGE-L} score on the \gls{CNNDM} corpus from 36.4 to 38.3 while being simpler to train. We also see comparable or better results than recent reinforcement-learning based methods with our MLE trained system. 
Furthermore, we find that the content selection model is very data-efficient and can be trained with less than 1\% of the original training data. This provides opportunities for domain-transfer and low-resource summarization. We show that a summarization model trained on \gls{CNNDM} and evaluated on the NYT corpus can be improved by over 5 points in \acrshort{ROUGE-L} with a content selector trained on only 1,000 in-domain sentences.  
 %Preliminary results further show potential of this technique for other domains such as data-to-text generation and grammar correction. 

\section{Related Work}
 
There is a tension in document summarization between staying close
to the source document and allowing compressive or abstractive modification. Many non-neural systems take a select and compress approach. For example, \citet{dorr2003hedge} introduced a system that first extracts noun and verb phrases from the first sentence of a news article and uses an iterative shortening algorithm to compress it. Recent systems such as \citet{durrett2016learning} also learn a model to select sentences and then compress them. 

In contrast, recent work in neural network based data-driven extractive summarization has focused on  extracting and ordering full sentences~\citep{cheng2016neural,dlikman2016using}. \citet{nallapati2016classify} use a classifier to determine whether to include a sentence and a selector that ranks the positively classified ones. 
%Many extraction methods use hidden markov models~\citep{conroy2001text} or conditional random fields~\citep{shen2007document} to predict the tags for the sentences. 
These methods often over-extract, but extraction at a word level requires maintaining grammatically correct output~\citep{cheng2016neural}, which is difficult. Interestingly, key phrase extraction while ungrammatical often matches closely in content with human-generated summaries~\citep{bui2016extractive}.
 
A third approach is neural abstractive summarization with
sequence-to-sequence models
\cite{sutskever2014sequence,bahdanau2014neural}. These methods have
been applied to tasks such as headline
generation~\citep{rush2015neural} and article
summarization~\citep{nallapati2016abstractive}. \citet{chopra2016abstractive}
show that attention approaches that are more specific to summarization
can further improve the performance of
models. \citet{gu2016incorporating} were the first to show that a copy
mechanism, introduced by \citet{vinyals2015pointer}, can combine the
advantages of both extractive and abstractive summarization by copying
words from the source. \citet{see2017get} refine this pointer-generator approach and use an
additional coverage mechanism \citep{tu2016modeling} that makes a
model aware of its attention history to prevent repeated attention.

Most recently, reinforcement learning (RL) approaches that optimize objectives for summarization other than maximum likelihood have been shown to further improve performance on these tasks \citep{paulus2017deep,li2018actor, celikyilmaz2018deep}. 
\citet{paulus2017deep} approach the coverage problem with an intra-attention in which a decoder has an attention over previously generated words. However RL-based training can be difficult to tune and slow to train.
Our method does not utilize RL training, although in theory this approach can be adapted to RL methods.

%and an inference penalty that prevents repeating trigrams. In this work, we extend the inference penalty work and show that summary-specific inference-time penalties can lead to summaries of similar quality as those that have a built-in coverage mechanism. %As another metric, \cite{fan2017controllable} show that by adding desired output features to the input of the encoder, a model can learn to generate text that has certain style and length properties.

Several papers also explore multi-pass extractive-abstractive summarization.
\citet{nallapati2017summarunner} create a new source document
comprised of the important sentences from the source and then train an
abstractive system. \citet{liu2018generating} describe an extractive
phase that extracts full paragraphs and an abstractive one that
determines their order.  Finally~\citet{zeng2016efficient} introduce a
mechanism that reads a source document in two passes and uses the
information from the first pass to bias the second. Our method differs in that we utilize a completely abstractive model, biased with a powerful
content selector.

Other recent work explores alternative approaches to content selection. For example, \citet{cohan2018discourse} use a hierarchical attention to detect relevant sections in a document, \citet{li2018guiding} generate a set of keywords that is used to guide the summarization process, and \citet{pasunuru2018multi} develop a loss-function based on whether salient keywords are included in a summary. Other approaches investigate the content-selection at the sentence-level. \citet{tan2017abstractive} describe a graph-based attention to attend to one sentence at a time, \citet{chen2018fast} first extract full sentences from a document and then compress them, and \citet{hsu2018unified} modulate the attention based on how likely a sentence is included in a summary.
 
\section{Background: Neural Summarization}

Throughout this paper, we consider a set of pairs of texts $(\mathcal{X}, \mathcal{Y})$ where $x \in \mathcal{X}$ corresponds to source tokens $x_1, \ldots, x_n$ and $y \in \mathcal{Y}$ to a summary $y_1, \ldots, y_m$ with $m \ll n$. 

 Abstractive summaries are generated one word at a time. At every time-step, a model is aware of the previously generated words. The problem is to learn a function $f(x)$ parametrized by $\theta$ that maximizes the probability of generating the correct sequences.
Following previous work, we model the abstractive summarization with an attentional sequence-to-sequence model. The attention distribution $p(a_j|x,y_{1:j-1})$ for a decoding step $j$, calculated within the neural network, represents an embedded soft distribution over all of the source tokens and can be interpreted as the current focus of the model. 

%We further use $c_j^* = \sum_i a_j^i c_i$ to refer to the context vector. Finally, let $r_{\textrm{out}} = W_{\textrm{out}}[h_j,c_j^*]+b_{\textrm{out}}$ be an intermediary representation derived from the context and decoder with $W_{\textrm{out}}$ and $b_{\textrm{out}}$ as trainable parameters.

The model additionally has a copy mechanism~\citep{vinyals2015pointer} to copy words from the source. Copy models extend the decoder by predicting a binary soft switch $z_j$ that determines whether the model copies or generates. The copy distribution is a probability distribution over the source text, and the joint distribution is computed as a convex combination of the two parts of the model,
\begin{align}
\label{eq:copy}
\begin{split}
& p(y_j\ |\  y_{1:j\text{-}1}, x) =  \\
&\  p(z_j~\textequal~1\ |\ y_{1:j\text{-}1}, x) \times p(y_j\ |\ z_j~\textequal~1, y_{1:j\text{-}1}, x) + \\
                        & \ p(z_j~\textequal~0\ |\ y_{1:j\text{-}1}, x) \times p(y_j\ |\ z_j~\textequal~0,y_{1:j\text{-}1}, x)
\end{split}
\end{align}
\noindent where the two parts represent copy and generation distribution respectively.
%Following \citet{gulcehre2016pointing}, we assume that every word that occurs in both source and target was copied to avoid having to marginalize over $z$, and maximize the joint likelihood of $y_j$ and $z_j$. 
%We define $z_j = \sigma(r_{\textrm{out}} v_{\textrm{copy}})$. 
Following the \emph{pointer-generator} model of \citet{see2017get}, we reuse the attention $p(a_j|x,y_{1:j-1})$ distribution as copy distribution, i.e. the copy probability of a token in the source $w$ through the copy attention is computed as the sum of attention towards all occurrences of $w$. During training, we maximize marginal likelihood with the latent switch variable. %, such that

% \begin{align*}
% p(y_j = w | z_j=1) = \sum_{i:w_i=w} a_j^i.
% \end{align*}

\begin{figure}[t]
\centering
\includegraphics[width=.45\textwidth]{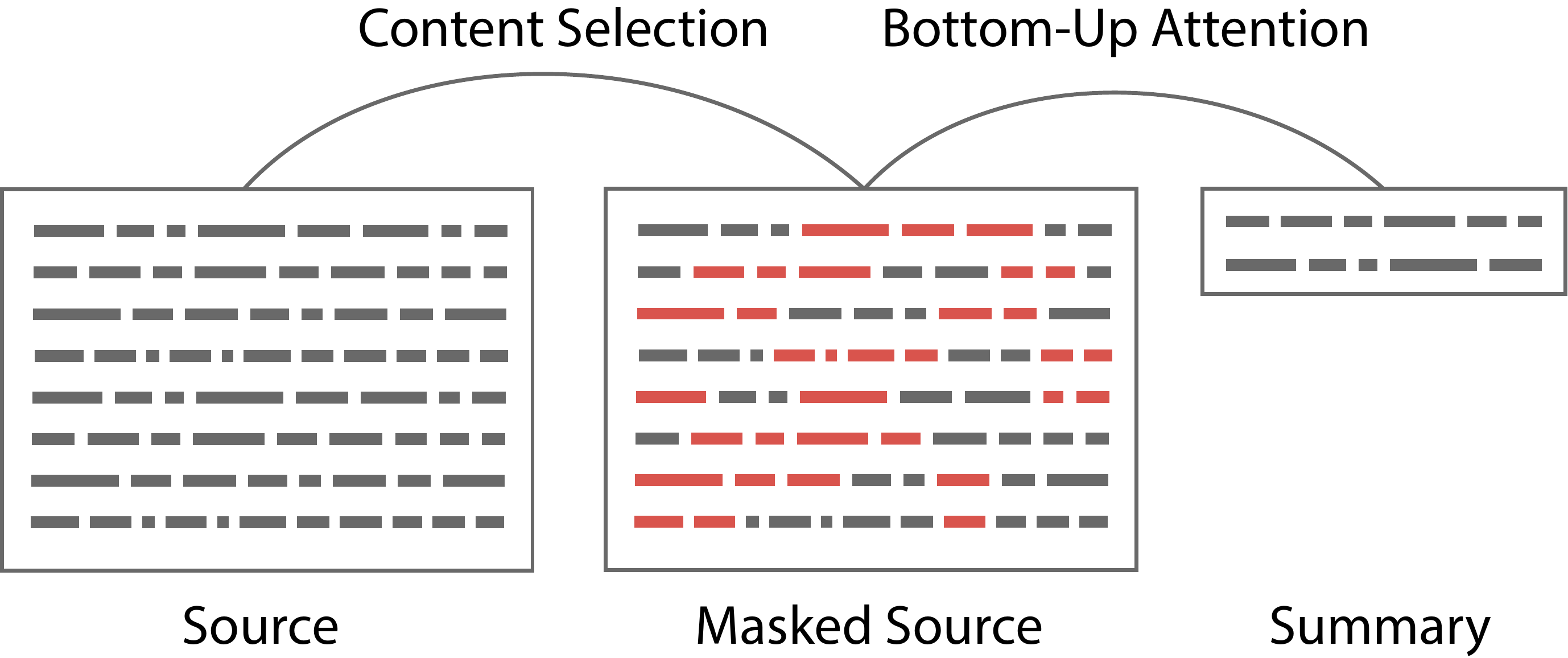}
\caption{Overview of the selection and generation processes described throughout Section~\ref{sec:bua}.}
\label{fig:bua}
\end{figure}

\section{Bottom-Up Attention}
\label{sec:bua}

We next consider techniques for incorporating a content selection into abstractive summarization, illustrated in Figure~\ref{fig:bua}. 

%Throughout, we refer to the detection of relevant phrases as content selection.

%, and the restriction of the copy attention to selected phrases as the bottom-up attention.
%Both techniques are only applied at inference-time, such that any trained summarization model can be extended to use them without any further training.  
\subsection{Content Selection}
\label{sec:content}

We define the content selection problem as a word-level extractive summarization task. While there has been significant work on custom extractive summarization (see related work), we make a simplifying assumption and treat it as a sequence tagging problem. Let $t_1, \ldots, t_n$ denote binary tags for each of the source tokens, i.e. $1$ if a word is copied in the target sequence and 0 otherwise. 

While there is no supervised data for this task, 
we can generate training data by aligning the summaries 
to the document. We define a word $x_i$ as copied if (1) it is part of the longest possible subsequence of tokens $s = x_{i-j:i:i+k}$, for integers $j \leq i; k \leq (n-i)$, if $s \in x $ and $s \in y$, and (2) there exists no earlier sequence $u$ with $s = u$.

We use a standard bidirectional LSTM model trained with maximum likelihood for the sequence labeling problem. Recent results have shown that better word representations can lead to significantly improved performance in sequence tagging tasks~\citep{peters2017semi}. Therefore, we first map each token $w_i$ into two embedding channels $e_i^{\textrm{(w)}}$ and $e_i^{\textrm{(c)}}$. The $e^{\textrm{(w)}}$ embedding represents a static channel of pre-trained word embeddings, e.g. GLoVE \citep{pennington2014glove}. The $e^{\textrm{(c)}}$ are contextual embeddings from a pretrained language model, e.g. ELMo  \citep{peters2018deep} %These contextualized word embeddings use a bidirectional language model that computes the probability of token $w_i$ given the preceding and succeeding tokens. % such that the probability of a sequence (omitting parameters for brevity) is modeled as
% \[
% \sum_{i=1}^n \left(p(w_i|w_{1:i-1}) + p(w_i|w_{i+1:n} \right).
% \]
which uses a character-aware token embedding~\citep{kim2016character} followed by two bidirectional LSTM layers $h^{(1)}_{i}$ and $h^{(2)}_{i}$. The contextual embeddings are fine-tuned to learn a task-specific embedding $e_i^{\textrm{(c)}}$ as a linear combination of the states of each LSTM layer and the token embedding,

\[
e^{\textrm{(c)}}_i = \gamma \times \sum_{\ell=0}^2 s_j \times h_{i}^{(\ell)}, 
\]

\noindent with $\gamma$ and $s_{0,1,2}$ as trainable parameters. Since these embeddings only add four additional parameters to the tagger, it remains very data-efficient despite the high-dimensional embedding space. 

Both embeddings are concatenated into a single vector that is used as input to a bidirectional LSTM, which computes a representation $h_i$ for a word $w_i$. We can then calculate the probability $q_i$ that the word is selected as $\sigma(W_sh_i + b_s)$ with trainable parameters $W_s$ and $b_s$.

\subsection{Bottom-Up Copy Attention}
\label{sec:rest}
Inspired by work in bottom-up attention for images~\citep{anderson2017bottom} which restricts attention to predetermined bounding boxes within an image, we use these attention masks to
limit the available selection of the pointer-generator model.

As shown in Figure~\ref{fig:ex}, a common mistake made by neural copy models is copying very long sequences or even whole sentences. In the baseline model, over 50\% of copied tokens are part of copy sequences that are longer than 10 tokens, whereas this number is only 10\% for reference summaries. While bottom-up attention could also be used to modify the source encoder representations, we found that a standard encoder over the full text was effective at aggregation and therefore limit the bottom-up step to attention masking.

%To not lose any contextual information of the source, we still use the all tokens in the encoder and standard attention, but restrict the copy attention of the model. 

Concretely, we first train a pointer-generator model on the full dataset as well as the content selector defined above.
At inference time, to generate the mask, the content selector computes selection probabilities $q_{1:n}$ for each token in a source document. %This computation is separate from the abstractive model, and can thus be computed in advance, or even with a model trained on a different data set. We analyze in Section~\ref{sec:eff}, how this property can be applied to domain transfer, and in Section~\ref{sec:ext} how the content selector can be applied as a fully extractive model. 
The selection probabilities are used to modify the copy attention distribution to only include tokens identified by the selector. 
Let $a_j^i$ denote the attention at decoding step $j$ to encoder word $i$.
Given a threshold $\epsilon$, the selection is applied as a hard mask, such that

\[
p(\tilde{a}_j^i|x, y_{1:j-1}) = 
\begin{cases}
p(a_j^i|x, y_{1:j-1}) &  q_{i} > \epsilon \\
0 & \textrm{ow. }
\end{cases}
\]

\noindent  To ensure that Eq.~\ref{eq:copy} still yields a correct probability distribution, we first multiply $p(\tilde{a}_j|x, y_{1:j-1})$ by a normalization parameter $\lambda$ and then renormalize the distribution. The resulting normalized distribution can be used to directly replace $a$ as the new copy probabilities.

% to ensure that Eq.~\ref{eq:copy} still yields a valid probability distribution. 
%The resulting normalized distribution can be used to directly replace $a$ as the new copy probability. 

%Let $e_j^i$ denote the unnormalized (log) attention score. The masked copy-attention $\tilde{a}_j^i$ of this step can be computed such that
%\[
%\tilde{a}_j^i = \frac{ \exp \lambda (e_j^i + \hat{t}_i)}{\sum_k \exp \lambda (e_j^k + \hat{t}_k)},
%\]
%\noindent with a tuning parameter $\lambda$. 
% This changes the probability for each word to

% \begin{align*}
% p(y_j = w | z_j=1) = \sum_{i:w_i=w} \tilde{a}_j^i.
% \end{align*}

%\noindent As a practical consideration, we found that setting a threshold such that approximately 25\% of the original text can be copied helpful for the combined approach, whereas a threshold that results in roughly 10\% of the original (i.e. 90\% compression) works better for a purely extractive approach. 

\subsection{End-to-End Alternatives}

Two-step \textsc{Bottom-Up} attention has the advantage of training simplicity. In theory, though, standard copy attention should be able to learn how to perform content selection as part of the end-to-end training. We consider several other end-to-end approaches for incorporating content selection into neural training.

% Incorporating task-specific modules directly into an end-to-end trained model has the advantage that it reduces potential errors that accumulate within a pipeline. Some modules, such as the copy attention, additionally improve the performance of a model in a way impossible for a pipelined approach. On the other hand, separate models can spend more resources on a single task, potentially leading to a better performance. 
% We compare both approaches by defining one two-step approach, and three end-to-end trained models.

%Since bottom-up attention typically describes a two-step approach, we call the pipeline approach the \textbf{Bottom-Up Attention}. In this approach, both models are trained separately, and we use the content selector to compute the masks. During inference of the abstractive model, we replace $a$ with $\tilde{a}$ as the copy attention, as described in Section~\ref{sec:rest}. 

Method 1: (\textsc{Mask Only}): We first consider whether the alignment used in the bottom-up approach could help a standard summarization system. Inspired by \citet{nallapati2017summarunner}, we investigate whether aligning the summary and the source during training and fixing the gold copy attention to pick the "correct" source word is beneficial. We can think of this approach as limiting the set of possible copies to a fixed source word. Here the training is changed, but no mask is used at test time.

% While the formulation of the copy attention avoids marginalization over all possible copy probabilities, the beam search inference is still searching over multiple paths and their associated copy probabilities. Applying bottom-up attention fixes some of the latent variables, which can reduce this complexity and therefore improve performance. To investigate whether a changed training regime can further take advantage of fixed latent variables, we define a baseline \textsc{Mask Only}, in which we apply the correct mask during training to guide the training of the copy attention. 

Method 2 (\textsc{Multi-Task}): Next, we investigate whether the content selector can be trained alongside the abstractive system. We first test this hypothesis by posing summarization as a multi-task problem and training the tagger and summarization model with the same features. For this setup, we use a shared encoder for both abstractive summarization and content selection. At test time, we apply the same masking method as bottom-up attention. 

Method 3 (\textsc{DiffMask}): Finally we consider training the full system end-to-end with the mask during training. Here we jointly optimize both objectives, but use predicted selection probabilities to softly mask the copy attention $p(\tilde{a}_j^i|x, y_{1:j-1}) = p(a_j^i|x, y_{1:j-1}) \times q_{i}$, which leads to a fully differentiable model. This model is used with the same soft mask at test time.

\section{Inference}

Several authors have noted that longer-form neural generation still has significant issues with incorrect length and repeated words than in short-form problems like translation. Proposed solutions include modifying models with extensions such as a coverage mechanism~\citep{tu2016modeling,see2017get} or intra-sentence attention~ \citep{cheng2016long,paulus2017deep}. We instead stick to the theme of modifying inference, and modify the scoring function to include a length penalty $lp$ and a coverage penalty $cp$, and is defined as $s(x,y) = \log p(y|x)/lp(x) + cp(x;y)$.

\textit{Length:} To encourage the generation of longer sequences, we apply length normalizations during beam search. We use the length penalty by \citet{wu2016google}, which is formulated as

\[
lp(y) = \frac{(5 + |y|)^\alpha}{(5+1)^\alpha},
\]

\noindent with a tunable parameter $\alpha$, where increasing $\alpha$ leads to longer summaries. We additionally set a minimum length based on the training data.

\textit{Repeats:} Copy models often repeatedly attend to the same source tokens, generating the same phrase multiple times. 
% To prevent this, \citet{see2017get} introduce a coverage mechanism~\citep{tu2016modeling} that extends the input to calculating $z_j$  by the sum of attentions in previous steps. However, this mechanism requires an additional training step, since a model is first trained to convergence without coverage, and then fine-tuned with it enabled. 
% Another approach by \citet{paulus2017deep} utilizes intra-attention~\citep{cheng2016long}. This attention mechanism allows a model to attend to previous decoding steps. This teaches the model more explicitly what it already generated. 
We introduce a new summary specific coverage penalty,
\[
cp(x;y) = \beta \left(-n + \sum_{i=1}^n \max \left(1.0, \sum_{j=1}^{m} a^j_i \right)\right).
\]

\noindent Intuitively, this penalty increases whenever the decoder directs more than 1.0 of total attention within a sequence towards a single encoded token. By selecting a sufficiently high $\beta$, this penalty blocks summaries whenever they would lead to repetitions.
Additionally, we follow \citep{paulus2017deep} and restrict the beam search to never repeat trigrams.

\section{Data and Experiments}

We evaluate our approach on the \gls{CNNDM} corpus~\citep{hermann2015teaching,nallapati2016abstractive}, and the NYT corpus~\citep{sandhaus2008new}, which are both standard corpora for news summarization. The summaries for the \gls{CNNDM} corpus are bullet points for the articles shown on their respective websites, whereas the NYT corpus contains summaries written by library scientists. \gls{CNNDM} summaries are full sentences, with on average 66 tokens ($\sigma=26$) and 4.9 bullet points. 
NYT summaries are not always complete sentences and are shorter, with on average 40 tokens ($\sigma=27$) and 1.9 bullet points. 
Following \citet{see2017get}, we use the non-anonymized version of the \gls{CNNDM} corpus and truncate source documents to 400 tokens and the target summaries to 100 tokens in training and validation sets. 
For experiments with the NYT corpus, we use the preprocessing described by~\citet{paulus2017deep}, and additionally remove author information and truncate source documents to 400 tokens instead of 800. These changes lead to an average of 326 tokens per article, a decrease from the 549 tokens with 800 token truncated articles.
The target (non-copy) vocabulary is limited to 50,000 tokens for all models. 

The content selection model uses pre-trained GloVe embeddings of size 100, and ELMo with size 1024. The bi-LSTM has two layers and a hidden size of 256. Dropout is set to 0.5, and the model is trained with Adagrad, an initial learning rate of 0.15, and an initial accumulator value of 0.1. We limit the number of training examples to 100,000 on either corpus, which 
only has a small impact on performance. For the jointly trained content selection models, we use the same configuration as the abstractive model.

\begin{table*}[t]
\centering
\rowcolors{2}{}{lightgray!10}
\begin{tabular}{@{}llll@{}}
\toprule
Method 				            & R-1 & R-2 & R-L \\ \midrule
% SummaRuNNer* \citep{nallapati2017summarunner}      & 39.6 & 16.2  & 35.3 \\

Pointer-Generator \citep{see2017get}           & 36.44 & 15.66 & 33.42 \\
Pointer-Generator + Coverage \citep{see2017get} & 39.53 & 17.28 & 36.38 \\
ML + Intra-Attention \citep{paulus2017deep}   & 38.30 & 14.81 & 35.49 \\ \midrule
ML + RL \citep{paulus2017deep}      & 39.87 & 15.82 & 36.90 \\
Saliency + Entailment reward \citep{pasunuru2018multi}  & 40.43 & 18.00 & 37.10 \\
Key information guide network \citep{li2018guiding}  & 38.95 & 17.12 & 35.68 \\
Inconsistency loss \citep{hsu2018unified}    & 40.68 & 17.97 & 37.13 \\
Sentence Rewriting \citep{chen2018fast}  & 40.88 & 17.80 & \textbf{38.54} \\\midrule
Pointer-Generator (our implementation)       & 36.25 & 16.17 & 33.41 \\
Pointer-Generator + Coverage Penalty & 39.12 & 17.35 & 36.12 \\  
CopyTransformer + Coverage Penalty & 39.25 & 17.54 & 36.45 \\  
%Transformer + Copy + Coverage Penalty* & 39.25 & 17.54 & 36.45 \\ 
Pointer-Generator + Mask Only                       & 37.70 & 15.63 & 35.49 \\
Pointer-Generator + Multi-Task                      & 37.67 & 15.59 & 35.47 \\
Pointer-Generator + DiffMask                       & 38.45 & 16.88 & 35.81 \\ 
Bottom-Up Summarization & \textbf{41.22} & \textbf{18.68} & 38.34 \\
Bottom-Up Summarization (CopyTransformer) & 40.96 & 18.38 &  38.16 \\ \bottomrule
\end{tabular}
\caption[Caption for LOF]{Results of abstractive summarizers on the \gls{CNNDM} dataset.\protect\footnotemark ~The first section shows encoder-decoder abstractive baselines trained with cross-entropy. The second section describes reinforcement-learning based approaches. The third section presents our baselines and the attention masking methods described in this work. }
\label{tab:abs}
\end{table*}

For the base model, we re-implemented the Pointer-Generator model as described by \citet{see2017get}. To have a comparable number of parameters to previous work, we use an encoder with 256 hidden states for both directions in the one-layer LSTM, and 512 for the one-layer decoder. The embedding size is set to 128. 
The model is trained with the same Adagrad configuration as the content selector. Additionally, the learning rate halves after each epoch once the validation perplexity does not decrease after an epoch. We do not use dropout and use gradient-clipping with a maximum norm of 2.
We found that increasing model size or using the Transformer~\citep{vaswani2017attention} can lead to slightly improved performance, but at the cost of increased training time and parameters. 
We report numbers of a Transformer with copy-attention, which we denote CopyTransformer. In this model, we randomly choose one of the attention-heads as the copy-distribution, and otherwise follow the parameters of the big Transformer by \citet{vaswani2017attention}.

All inference parameters are tuned on a 200 example subset of the validation set. Length penalty parameter $\alpha$ and copy mask $\epsilon$ differ across models, with $\alpha$ ranging from $0.6$ to $1.4$, and $\epsilon$ ranging from $0.1$ to $0.2$. The minimum length of the generated summary is set to 35 for \gls{CNNDM} and 6 for NYT. While the Pointer-Generator uses a beam size of 5 and does not improve with a larger beam, we found that bottom-up attention requires a larger beam size of 10. The coverage penalty parameter $\beta$ is set to 10, and the copy attention normalization parameter $\lambda$ to 2 for both approaches. 
We use AllenNLP~\citep{gardner2018allennlp} for the content selector, and OpenNMT-py for the abstractive models~\citep{klein2017opennmt}.\footnote{Code and reproduction instructions can be found at \url{https://github.com/sebastianGehrmann/bottom-up-summary}}.

\footnotetext{These results compare on the non-anonymized version of this corpus used by \citep{see2017get}. The best results on the anonymized version are  R1:41.69  R2:19.47  RL:37.92 from \citep{celikyilmaz2018deep}. We compare to their DCA model on the NYT corpus.}

\section{Results}

Table~\ref{tab:abs} shows our main results on the \gls{CNNDM} corpus, with abstractive models shown in the top, and bottom-up attention methods at the bottom. We first observe that using a coverage inference penalty scores the same as a full coverage mechanism, without requiring any additional model parameters or model fine-tuning. The results with the CopyTransformer and coverage penalty indicate a slight improvement across all three scores, but we observe no significant difference between Pointer-Generator and CopyTransformer with bottom-up attention. 

We found that none of our end-to-end models lead to improvements, indicating that it is difficult to apply the masking during training without hurting the training process. The \emph{Mask Only} model with increased supervision on the copy mechanism performs very similar to the \emph{Multi-Task} model. On the other hand, bottom-up attention leads to a major improvement across all three scores. While we would expect better content selection to primarily improve ROUGE-1, the fact all three increase hints that the fluency is not being hurt specifically. Our cross-entropy trained approach even outperforms all of the reinforcement-learning based approaches in ROUGE-1 and 2, while the highest reported ROUGE-L score by \citet{chen2018fast} falls within the 95\% confidence interval of our results.
 
Table~\ref{tab:nytres} shows experiments with the same systems on the NYT corpus. We see that
the 2 point improvement compared to the baseline Pointer-Generator maximum-likelihood approach carries 
over to this dataset. Here, the model outperforms the RL based model by \citet{paulus2017deep} in ROUGE-1 and 2, but not L, and is comparable to the results of \cite{celikyilmaz2018deep} except for ROUGE-L. The same can be observed when comparing ML and our Pointer-Generator. We suspect that a difference in summary lengths due to our inference parameter choices leads to this difference, but did not have access to their models or summaries to investigate this claim. This shows that a bottom-up approach achieves competitive results even to models that are trained on summary-specific objectives. 

% Please add the following required packages to your document preamble:
% \usepackage{booktabs}
\begin{table}[t]
\centering
\rowcolors{2}{}{lightgray!10}
\begin{tabular}{@{}llll@{}}
\toprule
Method & R-1 & R-2 & R-L \\ \midrule
\small{ML*} & 44.26 & 27.43 & 40.41 \\
\small{ML+RL*} & 47.03 & 30.72 & \textbf{43.10} \\
\small{DCA$^\dagger$} & \textbf{48.08} & 31.19 & 42.33 \\
\small{Point.Gen. + Coverage Pen.} & 45.13 & 30.13 & 39.67 \\
\small{Bottom-Up Summarization} & 47.38 & \textbf{31.23} & 41.81 \\ \bottomrule
\end{tabular}
\caption{Results on the NYT corpus, where we compare to RL trained models. * marks models and results by \citet{paulus2017deep}, and $^\dagger$ results by \citet{celikyilmaz2018deep}. }
\label{tab:nytres}
\end{table}

% As mentioned before, we selected the baselines to investigate whether providing supervision for latent variables during training or inference lead to better summaries, and whether separate sequence tagging models are more data efficient. The results indicate that fixing latent variables during training does not lead to an improvement in this case, whereas it does during inference. However, this improvement is only observable with a separate tagging model, and not with an end-to-end trained one. 
The main benefit of bottom-up summarization seems to be from the reduction of mistakenly copied words. With the best Pointer-Generator models, the precision of copied words is 50.0\%  compared to the reference. This precision increases to 52.8\%, which mostly drives the increase in R1. An independent-samples t-test shows that this improvement is statistically significant with $t\text{=}14.7$ ($p<10^{-5}$). We also observe a decrease in average sentence length of summaries from 13 to 12 words when adding content selection compared to the Pointer-Generator while holding all other inference parameters constant. %We are further exploring the data efficiency in Section~\ref{sec:eff}.

\begin{figure}[t]
\centering
\includegraphics[width=.45\textwidth]{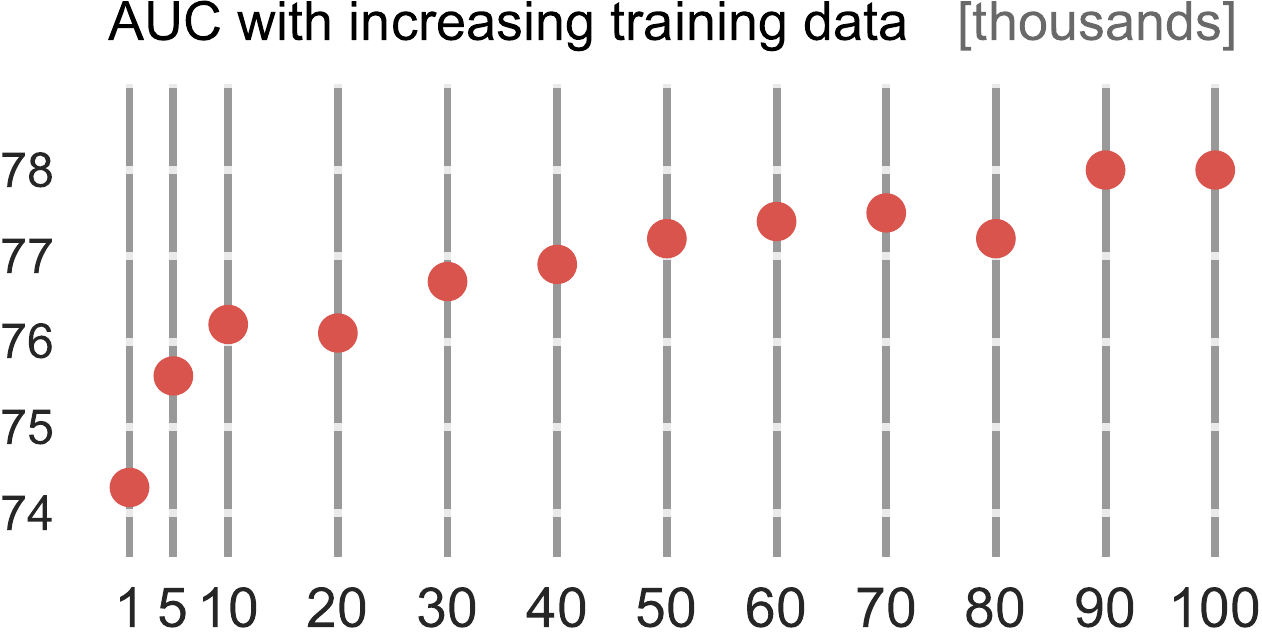}
\caption{The AUC of the content selector trained on \gls{CNNDM} with different training set sizes ranging from 1,000 to 100,000 data points.}
\label{fig:auc_increase}
\end{figure}

\paragraph{Domain Transfer}
While end-to-end training has become common, there are benefits to a two-step method.
Since the content selector only needs to solve a binary tagging problem with pretrained vectors, it performs well even with very limited training data. As shown in Figure~\ref{fig:auc_increase}, with only 1,000 sentences, the model achieves an AUC of over 74. Beyond that size, the AUC of the model increases only slightly with increasing training data.

To further evaluate the content selection, we consider an application to domain transfer. In this experiment, we apply the Pointer-Generator trained on \gls{CNNDM} to the NYT corpus. In addition, we train three content selectors on 1, 10, and 100 thousand sentences of the NYT set, and use these in the bottom-up summarization. 
The results, shown in Table~\ref{tab:nyt}, demonstrates that even a model trained on the smallest subset leads to an improvement of almost 5 points over the model without bottom-up attention. This improvement increases with the larger subsets to up to 7 points. While this approach does not reach a comparable performance to models trained directly on the NYT dataset, it still represents a significant increase over the not-augmented \gls{CNNDM} model and produces summaries that are quite readable. We show two example summaries in Appendix~\ref{app:transfer-examples}. This technique could be used for low-resource domains and for problems with limited data availability. 

% Please add the following required packages to your document preamble:
% \usepackage{booktabs}
\begin{table}[t]
\centering
\rowcolors{2}{}{lightgray!10}
\begin{tabular}{@{}lllll@{}}
\toprule
 & AUC & R-1 & R-2 & R-L \\ \midrule
CNNDM &  & 25.63 & 11.40 & 20.55 \\
+1k & 80.7 & 30.62 & 16.10 & 25.32 \\
+10k & 83.6 & 32.07 & 17.60 & 26.75 \\
+100k & 86.6 & 33.11 & 18.57 & 27.69 \\ \bottomrule
\end{tabular}
\caption{Results of the domain transfer experiment. AUC numbers are shown for content selectors. ROUGE scores represent an abstractive model trained on \gls{CNNDM} and evaluated on NYT, with additional copy constraints trained on 1/10/100k training examples of the NYT corpus. }
\label{tab:nyt}
\end{table}

\section{Analysis and Discussion}

\begin{table}[t]
\centering
\rowcolors{2}{}{lightgray!10}
\begin{tabular}{@{}lrrrr@{}}
\toprule
Method                & R-1   & R-2  & R-L  \\ \midrule
LEAD-3                & 40.1  & 17.5 & 36.3 \\
\small{NEUSUM} \citep{zhou2018neural} & 41.6  & 19.0 & 38.0 \\
Top-3 sents (Cont. Select.)    & 40.7  & \textbf{18.0} & 37.0 \\ \midrule
%Sents avg $>0.3$ &      & 39.6  & 17.8 & 36.1 \\  
Oracle Phrase-Selector       & 67.2  & 37.8 & 58.2 \\
Content Selector             & \textbf{42.0}  & 15.9 & \textbf{37.3} \\ \bottomrule
%Phrase, repl.            & 78.4 & 41.8  & 16.4 & 37.1 \\
%Phrase, cov. 0.8         & 78.0 & 41.9  & 16.8 & 37.2 \\ 
% Phrase (processed)  & 78.2 & 42.1  & 17.0 & 37.6 \\ \bottomrule
\end{tabular}
\caption{Results of extractive approaches on the CNN-DM dataset. The first section shows sentence-extractive scores. The second section first shows an oracle score if the content selector selected all the correct words according to our matching heuristic. Finally, we show results when the Content Selector extracts all phrases above a selection probability threshold.}
\label{tab:ext}
\end{table}

\paragraph{Extractive Summary by Content Selection?}
\label{sec:ext}
Given that the content selector is effective in conjunction with the abstractive model, it is interesting to know whether it has learned an effective extractive summarization system on its
own. Table~\ref{tab:ext} shows experiments comparing content selection to extractive baselines. 
The LEAD-3 baseline is a commonly used baseline in news summarization that extracts the first three sentences from an article. %Since news articles are written such that the beginning of an article summarizes the content, it provides a very strong baseline, sometimes even outperforming abstractive models. 
Top-3 shows the performance when we extract the top three sentences by average copy probability from the selector. Interestingly, with this method, only 7.1\% of the top three sentences are not within the first three, further reinforcing the strength of the LEAD-3 baseline. Our naive sentence-extractor performs slightly worse than the highest reported extractive score by \citet{zhou2018neural} that is specifically trained to score combinations of sentences.  
The final entry shows the performance when all the words above a threshold are extracted such that the resulting summaries are approximately the length of reference summaries. The oracle score represents the results if our model had a perfect accuracy, and shows that the content selector, while yielding competitive results, has room for further improvements in future work. 

This result shows that the model is quite effective at finding important words (ROUGE-1) but less effective at chaining them together (ROUGE-2). Similar to \citet{paulus2017deep}, we find that the decrease in ROUGE-2 indicates a lack of fluency and grammaticality of the generated summaries. A typical example looks like this:

\begin{quote}
a man food his first hamburger wrongfully for 36 years. michael hanline, 69, was convicted of murder for the shooting of truck driver jt mcgarry in 1980 on judge charges.
\end{quote}

\noindent This particular ungrammatical example has a ROUGE-1 of 29.3. This further highlights the benefit of the combined approach where bottom-up predictions are chained together fluently by the abstractive system. However, we also note that the abstractive system requires access to the full source document. Distillation experiments in which we tried to use the output of the content-selection as training-input to abstractive models showed a drastic decrease in model performance.

\begin{table}[t]
\centering
\rowcolors{2}{}{lightgray!10}
\begin{tabular}{@{}lrrrr@{}}
\toprule
% Data         & \multicolumn{2}{l}{\% Novel} &VERB&NOUN&ADJ\\ \midrule
% Reference         & 14.8  & \scalerel*{\includegraphics{figs/15bar}}{B}  &30.9&35.5&12.3   \\
% Vanilla S2S         & 6.6  & \scalerel*{\includegraphics{figs/6_5bar}}{B}  &14.5&19.7&5.1   \\
% Copy + Cov  & 2.2  & \scalerel*{\includegraphics{figs/2bar}}{B}    &25.7&39.3&13.9          \\ 
% Copy, Cov, Ext    & 0.5  & \scalerel*{\includegraphics{figs/0_5bar}}{B}  &53.3&24.8&6.5           \\ \bottomrule
Data                 & \%Novel &Verb&Noun&Adj\\ \midrule
\small{Reference}            & 14.8 &30.9&35.5&12.3  \\
\small{Vanilla S2S}         & 6.6  &14.5&19.7&5.1   \\
\small{Pointer-Generator}   & 2.2  &25.7&39.3&13.9  \\ 
\small{Bottom-Up Attention} & 0.5  &53.3&24.8&6.5   \\ \bottomrule
\end{tabular}
\caption{\%Novel shows the percentage of words in a summary that are not in the source document. The last three columns show the part-of-speech tag distribution of the novel words in generated summaries.}
\label{tab:copy}
\end{table}
%\newpage

\paragraph{Analysis of Copying}

While Pointer-Generator models have the ability to abstract in summary, the use of a copy mechanism 
causes the summaries to be mostly extractive. Table~\ref{tab:copy} shows that with copying the percentage of generated words that are not in the source document decreases from 6.6\% to 2.2\%, while reference summaries are much more abstractive with 14.8\% novel words. Bottom-up attention leads to a further reduction to only a half percent. However, since generated summaries are typically not longer than 40-50 words, the difference between an abstractive system with and without bottom-up attention is less than one novel word per summary. This shows that the benefit of abstractive models has been less in their ability to produce better paraphrasing but more in the ability to create fluent summaries from a mostly extractive process. 

Table~\ref{tab:copy} also shows the part-of-speech-tags of the novel generated words, and we can observe an interesting effect. Application of bottom-up attention leads to a sharp decrease in novel adjectives and nouns, whereas the fraction of novel words that are verbs sharply increases. When looking at the novel verbs that are being generated, we notice a very high percentage of tense or number changes, indicated by variation of the word ``say'', for example ``said'' or ``says'', while novel nouns are mostly morphological variants of words in the source. 

\begin{figure}[t]
\centering
\includegraphics[width=.45\textwidth]{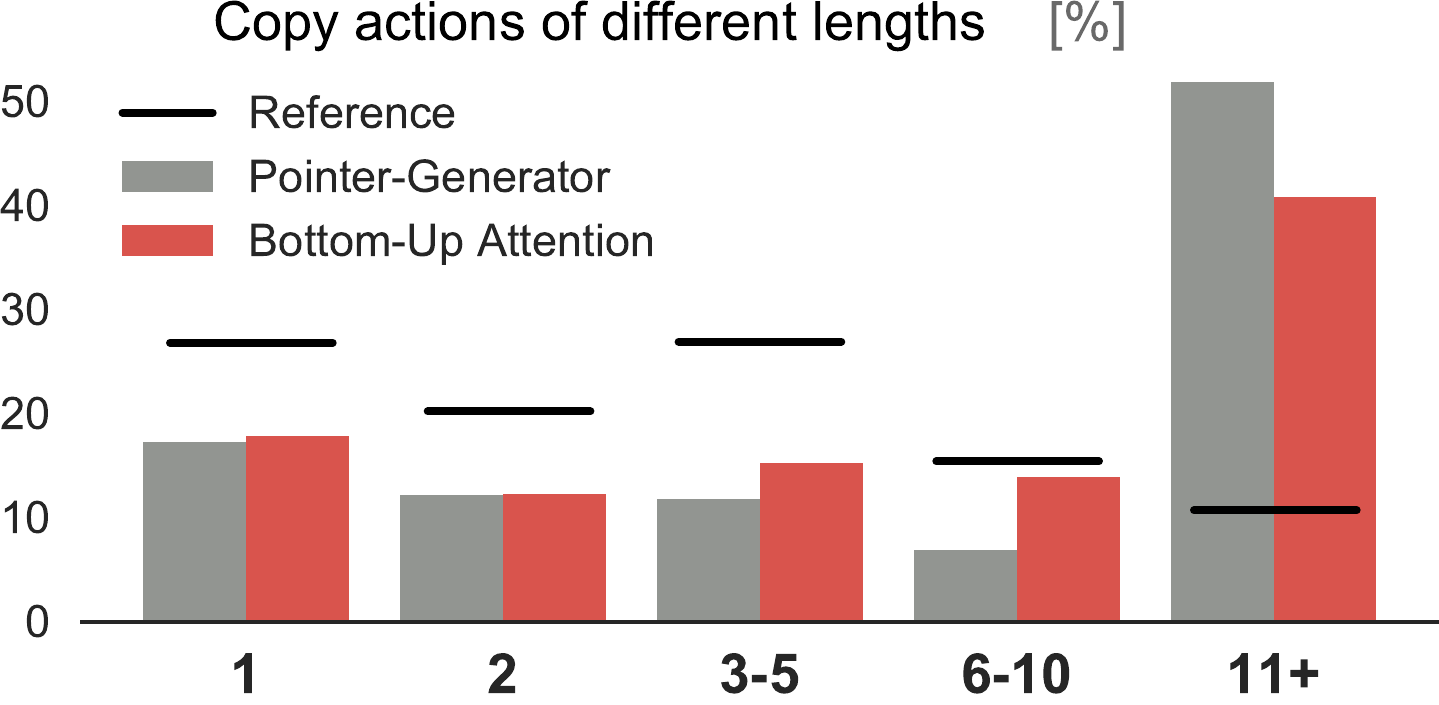}
\caption{For all copied words, we show the distribution over the length of copied phrases they are part of. The black lines indicate the reference summaries, and the bars the summaries with and without bottom-up attention.}
\label{fig:copy_length}
\end{figure}

Figure~\ref{fig:copy_length} shows the length of the phrases that are being copied. While most copied phrases in the reference summaries are in groups of 1 to 5 words, the Pointer-Generator copies many very long sequences and full sentences of over 11 words. Since the content selection mask interrupts most long copy sequences, the model has to either generate the unselected words using only the generation probability or use a different word instead. While we observed both cases quite frequently in generated summaries, the fraction of very long copied phrases decreases. However, either with or without bottom-up attention, the distribution of the length of copied phrases is still quite different from the reference.

% \subsection{Does this transfer to other domains?}

% \begin{figure}[t]
% \centering
% \includegraphics[width=.45\textwidth]{figs/position}
% \caption{Positions of actually copied tokens and predicted copied tokens. Both exhibit noticeable increases whenever a new sentence is started. The predicted phrases loosely follow the bumps of the actual data.}
% \label{fig:auc_increase}
% \end{figure}

\paragraph{Inference Penalty Analysis}

We next analyze the effect of the inference-time loss functions. Table~\ref{tab:inference-pen} presents the marginal improvements over the simple Pointer-Generator when adding one penalty at a time. We observe that all three penalties improve all three scores, even when added on top of the other two. This further indicates that the unmodified Pointer-Generator model has already learned an appropriate representation of the abstractive summarization problem, but is limited by its ineffective content selection and inference methods.

\section{Conclusion}
This work presents a simple but accurate content selection model for summarization that identifies phrases within a document that are likely included in its summary. We showed that this content selector can be used for a bottom-up attention that restricts the ability of abstractive summarizers to copy words from the source. The combined bottom-up summarization system leads to improvements in ROUGE scores of over two points on both the \gls{CNNDM} and NYT corpora. A comparison to end-to-end trained methods showed that this particular problem cannot be easily solved with a single model, but instead requires fine-tuned inference restrictions. Finally, we showed that this technique, due to its data-efficiency, can be used to adjust a trained model with few data points, making it easy to transfer to a new domain. Preliminary work that investigates similar bottom-up approaches in other domains that require a content selection, such as grammar correction, or data-to-text generation, have shown some promise and will be investigated in future work.

\begin{table}[t]
\centering
\rowcolors{2}{}{lightgray!10}
\begin{tabular}{@{}lrrr@{}}
\toprule
Data                        & R-1  &R-2  &R-L\\ \midrule
\small{Pointer Generator}   & 36.3 &16.2 &33.4  \\
\small{+ Length Penalty}    & 38.0 &16.8 &35.0   \\
\small{+ Coverage Penalty}  & 38.9 &17.2 &35.9  \\ 
\small{+ Trigram Repeat}    & 39.1 &17.4 &36.1   \\ \bottomrule
\end{tabular}
\caption{Results on \gls{CNNDM} when adding one inference penalty at a time.}
\label{tab:inference-pen}
\end{table}

\section*{Acknowledgements}

We would like to thank Barbara J. Grosz for helpful discussions and feedback on early stages of this work. We further thank the three anonymous reviewers. This work was supported by a Samsung Research Award. YD was funded in part by a Bloomberg Research Award. SG was funded in part by NIH grant 5R01CA204585-02. 

%\nocite{*}
\bibliography{emnlp2018}
\bibliographystyle{acl_natbib_nourl}

\begin{table*}[!ht]
\begin{tabular}{@{}lp{.7\textwidth}@{}}
\toprule
Examples & Generated summary \\ \midrule
Reference & green bay packers successful season is largely due to quarterback brett favre \\
S2S & ahman green rushed for 000 yards in 00-00 victory over the giants . true , dorsey levens , good enough to start for most teams but now green 's backup , contributed kickoff returns of 00 , 00 and 00 yards . \\
Content Selection & playoff-bound green bay packers beat the giants in the 00-00 victory . the packers won three games and six of each other . \\ \midrule
Reference & paul byers , pioneer of visual anthropology , dies at age 00 \\
S2S & paul byers , an early practitioner of mead , died on dec. 00 at his home in manhattan .  he enlisted in the navy , which trained him as a cryptanalyst and stationed him in australia . \\
Content Selection & paul byers , an early practitioner of anthropology , pioneered with margaret mead . \\ \bottomrule
\end{tabular}
\caption{\label{fig:transfer-examples}Domain-transfer examples.}
\end{table*}
% \newpage
\FloatBarrier
\appendix

\section{Domain Transfer Examples}
\label{app:transfer-examples}

We present two generated summaries for the \gls{CNNDM} to NYT domain transfer experiment in Table~\ref{fig:transfer-examples}. S2S refers to a Pointer-Generator with Coverage Penalty trained on \gls{CNNDM} that scores 20.6 ROUGE-L on the NYT dataset. The content-selection improves this to 27.7 ROUGE-L without any fine-tuning of the S2S model.

\end{document}